\documentclass{article} 
\usepackage{iclr2025_conference,times}

\usepackage{amsmath,amsfonts,bm}

\usepackage[most]{tcolorbox}

\def\Figref#1{Figure~\ref{#1}}

\def\Secref#1{Section~\ref{#1}}

\def\eqref#1{equation~\ref{#1}}

\def\1{\bm{1}}

\DeclareMathAlphabet{\mathsfit}{\encodingdefault}{\sfdefault}{m}{sl}
\SetMathAlphabet{\mathsfit}{bold}{\encodingdefault}{\sfdefault}{bx}{n}

\newtcolorbox[auto counter]{takeaway}[1][]{
    enhanced,
    breakable,
    colframe=lightblue,
    colback=lightblue!30!white,
    sharp corners,
    boxsep=0pt,
    left=5pt,
    right=5pt,
    top=6pt,
    bottom=6pt,
    boxrule=0pt,
    leftrule=4pt,
    before upper={\textbf{Takeaway \thetcbcounter: } },
    #1
}
\usepackage{pifont}

\newcommand{\greenyes}{\textcolor{green}{\ding{51}}}
\newcommand{\redno}{\textcolor{red}{\ding{55}}}

\usepackage{hyperref}
\usepackage{url}
\usepackage{booktabs}
\usepackage{multirow} 
\usepackage{subcaption}
\usepackage{arydshln}
\usepackage{todonotes}
\usepackage{utfsym}
\usepackage{algorithm}
\usepackage{algpseudocode}
\usepackage{amsmath}
\usepackage{setspace}
\usepackage{wrapfig} 
\usepackage{colortbl}
\usepackage{makecell}
\usepackage{tcolorbox}
\usepackage{diagbox}
\usepackage{arydshln}
\usepackage{enumitem}

\tcbuselibrary{most}
\definecolor{lightgreen}{RGB}{200,255,200}
\definecolor{lightpink}{rgb}{1.0, 0.85, 0.9} 
\definecolor{lightblue}{rgb}{0.529, 0.808, 0.922} 
\definecolor{lightgray}{gray}{0.85}

\newtcolorbox{mybox}[2][]
  {colback = black!5!white, colframe = black!75!black, fonttitle = \bfseries,
    colbacktitle = black!100!black, enhanced,
    attach boxed title to top left={yshift=-2.2mm,xshift=4mm},
    title=#2,#1}

\title{Scaling Up, Speeding Up: A Benchmark of Speculative Decoding for Efficient LLM Test-Time Scaling}

\author{Shengyin Sun$^{1,*}$, Yiming Li$^{2,*}$, Xing Li$^{2}$, Yingzhao Lian$^{2}$, Weizhe Lin$^{2}$, Hui-Ling Zhen$^{2}$,\\ 
\textbf{Zhiyuan Yang$^{2}$, Chen Chen$^{2}$, Xianzhi Yu$^{2}$, Mingxuan Yuan$^{2}$, Chen Ma$^{1,\dagger}$}\\
$^{1}$City University of Hong Kong, $^{2}$Huawei Noah’s Ark Lab\\
\texttt{shengysun4-c@my.cityu.edu.hk, li.yiming3@huawei.com}\\
\texttt{Yuan.Mingxuan@huawei.com, chenma@cityu.edu.hk}
}

\iclrfinalcopy 
\begin{document}

\maketitle

\begin{figure}[h]
    \centering
    \includegraphics[width=0.9\linewidth]{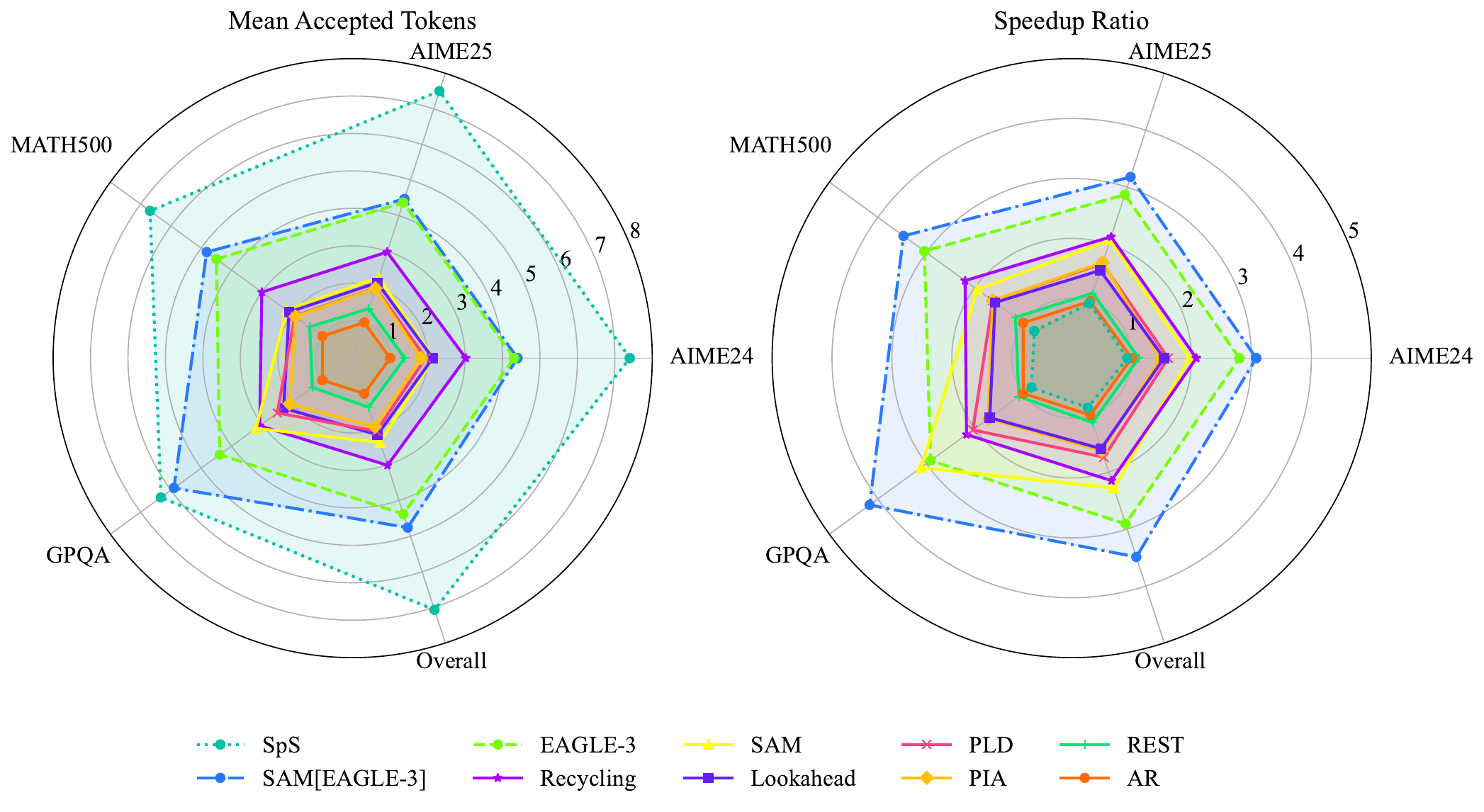}
    \caption{Performance of various model speculative decoding methods on Qwen3-8B under the multi-turn thinking framework with temperature $T=0$.}
    \label{ssy0801:radar_fig}
\end{figure}

{
\let\thefootnote\relax\footnotetext{Work in progress.}
\let\thefootnote\relax\footnotetext{$^*$ Equal contribution. $^\dagger$ Corresponding author.}
}

\textbf{(TL;DR) Summary of our findings:}
\begin{itemize}
    \item While the method relies on training an auxiliary draft model demonstrates promising acceleration potential, its performance is \textbf{inherently tied to the training process}, raising practical concerns about their adaptability to diverse reasoning scenarios.
    \item Methods leveraging N-gram patterns excel at capturing redundancy within test-time scaling, yet this advantage is accompanied by \textbf{sensitivity to sampling temperature}.
    \item Hybrid speculative decoding uniquely \textbf{combines} methods that train an auxiliary draft model for \textbf{semantic alignment} with those leveraging \textbf{N-gram patterns to capture repetition}, unlocking distinct potential for reasoning under test-time scaling.
    \item Methods that store N-gram patterns during generation can achieve progressive acceleration by reusing past computations, \textbf{enabling efficiency that scales across turns}.
    \item In hybrid speculative decoding, the acceleration from reusing repetitive token sequences is \textbf{limited by suboptimal integration strategies}, calling for more dynamic approaches to realize its potential under test-time scaling.
    \item Methods leveraging N-gram patterns incur \textbf{lower draft generation time overhead}, enabling a greater share of computational resources to be allocated to the decoding phase.
\end{itemize}

\clearpage

\begin{abstract}
Test-time scaling has emerged as a powerful paradigm for enhancing the reasoning capabilities of large language models (LLMs) by allocating additional computational resources during inference. However, this paradigm is inherently inefficient due to the generation of redundant and repetitive reasoning traces, leading to significant computational overhead. Speculative decoding offers a promising avenue for mitigating this inefficiency, yet its efficacy in the structured, repetition-rich context of test-time scaling remains largely unexplored. To bridge this gap, we introduce the first comprehensive benchmark designed to evaluate speculative decoding methods for accelerating LLM test-time scaling. Our benchmark provides consistent experimental protocols across representative test-time scaling paradigms (e.g., Best-of-N sampling and multi-round thinking), enabling a fair comparison of three major categories of speculative decoding: model-based, training-based, and n-gram-based methods. Extensive experiments reveal that simple n-gram-based methods effectively capture repetitive patterns, demonstrating unique potential in accelerating test-time scaling. This phenomenon demonstrates the value of integrating n-gram-based methods with model-based or training-based approaches to balance acceleration for both repetitive and diverse reasoning in test-time scaling. We hope this benchmark spurs further research on speculative decoding for test-time scaling, enabling faster and more practical reasoning in LLMs through better handling of repetitive and diverse reasoning paths.
\end{abstract}

\section{Introduction}
\label{sec:intro}

In the pursuit of more capable Large Language Models (LLMs), a powerful paradigm known as test-time scaling~\citep{DBLP:conf/iclr/snelllee25,DBLP:conf/iclr/wangwei23,DBLP:conf/nips/madaantandon23,DBLP:arxiv/tianzhao25,DBLP:journals/nips/yaoyu23} has emerged as a key driver of performance on complex tasks. This approach enhances model performance not by altering the model itself, but by allocating substantial computational resources during the inference phase to allow the model to ``think" longer and more deeply about a given problem. Best-of-N sampling is a prominent method in this paradigm, which generates multiple candidate solutions and selects the best one~\citep{DBLP:conf/iclr/snelllee25}. Another widely adopted approach is multi-round thinking, which employs iterative refinement to progressively improve the model’s answer over several turns~\citep{DBLP:arxiv/tianzhao25}. These methods have demonstrated remarkable success in complex reasoning~\citep{DBLP:arxiv/lightmankosaraju23}, mathematics~\citep{DBLP:arxiv/cobbekosaraju21}, and coding~\citep{DBLP:arxiv/lichoi22}. However, this enhanced performance comes at a steep price. Generating multiple full-length responses or iterative reasoning chains incurs prohibitive computational overhead, leading to a significant latency bottleneck that severely limits their use in real-time, interactive scenarios.

To unlock the potential of test-time scaling for practical use, it is imperative to mitigate its associated latency. Speculative decoding~\citep{DBLP:conf/acl/xiayang24,DBLP:conf/icml/leviathankalman23,DBLP:arxiv/chenborgeaud23} has emerged as a leading technique to achieve this goal. This approach accelerates autoregressive generation by using a fast draft mechanism to propose candidate token sequences, which are then efficiently verified in a single pass by the larger target model. This process leads to substantial reductions in wall-clock time while crucially guaranteeing a final output distribution that is mathematically identical to that of the original target model. This property makes speculative decoding an ideal candidate for acceleration, offering the promise of significant speedups without compromising the quality gains that make test-time scaling methods so attractive.

In turn, the repetitive and computationally intensive nature of test-time scaling creates uniquely favorable conditions for speculative decoding to excel, particularly for retrieval-based solutions~\citep{DBLP:conf/acl/huwang25,DBLP:arxiv/oliarojia24}. These retrieval-based approaches are inspired by N-gram models, which predict the next token based on a specific sequence of prior tokens, such as pairs (bigrams) or triples (trigrams). As a model explores different reasoning paths in Best-of-N or refines its thoughts in multi-round dialogue, it frequently outputs highly repetitive token sequences. These can include recurring logical phrases (e.g., ``Let's consider the case where..."), standard boilerplate code, or common transitional expressions that appear across multiple attempts (specific examples are provided in Table \ref{ssy0801:redundancy_example}). This repetitive structure suggests that N-gram-based speculative decoding methods~\citep{DBLP:conf/acl/huwang25,DBLP:arxiv/oliarojia24}, which dynamically cache and reuse recently generated token sequences, are exceptionally well-suited to this scenario. Their ability to anticipate and complete these common thought patterns offers a direct path to acceleration.

Despite the promise of acceleration, the existing literature lacks a comparative analysis to determine which speculative decoding architecture is best suited to exploit the computational redundancies inherent in test-time scaling. Current benchmarks typically evaluate these methods in generic, single-pass generation tasks~\citep{DBLP:conf/acl/xiayang24}, overlooking the unique characteristics of advanced reasoning paradigms. Consequently, critical questions remain unanswered: \textit{Can a simple, adaptive N-gram mechanism outperform a more complex, pre-trained draft model in a scenario ripe with repetition? How do the trade-offs between these methods—flexibility, training cost, and real-time adaptability—manifest when accelerating test-time scaling frameworks?} Without a dedicated benchmark to investigate these specific interactions, the community cannot make informed decisions about how to efficiently scale up inference-time reasoning.

\begin{table}[t]
  \centering
  \small
  \caption{Illustration of intra-turn and inter-turn redundancy under the test-time scaling paradigm.}
  \scalebox{0.87}{
    \begin{tabular}{ll} 
    \toprule[1.5pt]
    \makecell[l]{Question} & \multicolumn{1}{l}{\makecell[l]{%
    Let $p$ be the least prime number for which there exists a positive integer $n$ such that $n^{4}+1$ is divi-\\
    sible by $p^{2}$. Find the least positive integer $m$ such that $m^{4}+1$ is divisible by $p^{2}$.
}} \\
    \midrule
    \makecell[l]{Intra-turn\\Redundancy} & \makecell[l]{\textbf{Turn 1:} \colorbox{yellow!30}{Compute $8^3 = 512$. Then $512 \bmod 17$: since $17 \times 30 = 510$, we have $512 \equiv 2 \pmod{17}$.}\\ {Therefore, $4 \times 8^3 \equiv 4 \times 2 = 8 \pmod{17}$.} Thus, $4 \times 8^3 \times 17k \equiv 8 \times 17k \pmod{289}$. So, $n^4 \equiv $\\$50 + 8 \times 17k \pmod{289}$. Hence, $n^4 + 1 \equiv 51 + 136k \pmod{289}$. We need this to be $0$ $(\rm{mod}$\\$289)$, so $51 + 136k \equiv 0 \pmod{289}$. Dividing both sides by $17$, we get $3 + 8k \equiv 0 \pmod{17}$, i.e., \\$\cdots$ Wait, earlier I said that $4\times8^3\times17k \equiv 8\times17k$ mod 289. Let me check that again. $4\times8^3.$ \\\colorbox{yellow!30}{Compute $8^3 = 512$. Then $512 \bmod 17$: since $17 \times 30 = 510$, we} \colorbox{yellow!30}{have $512 \equiv 2 \pmod{17}$.}}\\ 
    
    \makecell[l]{Inter-turn\\Redundancy} & \multicolumn{1}{l}{\makecell[l]{\textbf{Turn 1:} \textcolor{blue}{\texttt{$<$think$>$}}\textcolor{red}{\texttt{\textbackslash n}} \colorbox{green!30}{Okay, so I need to find the least prime} number\colorbox{green!30}{ p such that there's a posit-}\\
    \colorbox{green!30}{ive integer n where $p^2$ divides $n^4 + 1$. Then, once I find that p, I have to find the smallest positive}\\
    \colorbox{green!30}{integer m such that $p^2$ divides $m^4 + 1$}. Alright, let me start by understanding the problem $\cdots$\\
    \textbf{Turn 2:} \textcolor{blue}{\texttt{$<$think$>$}}\textcolor{red}{\texttt{\textbackslash n}}\colorbox{green!30}{Okay, so I need to find the least prime p such that there's a positive integer}\\ 
    \colorbox{green!30}{$n$ with $p^2$ dividing $n^4+1$. Then, once I find that $p$, I need to find the smallest m such that $m^4 + 1$}\\
    \colorbox{green!30}{is divisible by $p^2$}. \textcolor{red}{\texttt{\textbackslash n}}\textcolor{red}{\texttt{\textbackslash n}}First, let me recall that$\cdots$
    }}\\
    \bottomrule[1.5pt]
    \end{tabular}}
  \label{ssy0801:redundancy_example}%
\end{table}

To address this critical gap, this paper introduces the first comprehensive benchmark designed to systematically compare different families of speculative decoding methods for accelerating LLM test-time scaling. Specifically, our work provides a head-to-head comparison of 
9 speculative decoding methods (more details can be seen in Section \ref{ssy0801:diff_methods}), covering nearly all SOTA techniques in this domain that are compatible with reasoning models. We aim to elucidate the trade-offs among these approaches and, most importantly, test our hypothesis on the unique effectiveness of using N-gram patterns to accelerate LLM test-time scaling. Our main observations are summarized as follows: (i) The method that relies on training an auxiliary draft model demonstrate promising acceleration potential, but its inference behavior is inherently coupled to the training process, limiting adaptability across diverse reasoning scenarios. (ii) Methods leveraging N-gram patterns excel at capturing redundancy in test-time scaling; in particular, those that store and reuse N-gram patterns during generation achieve progressive acceleration by reusing past computations, enabling efficiency gains that accumulate across generation steps. However, this advantage is sensitive to sampling temperature. (iii) Hybrid speculative decoding uniquely combines methods that train an auxiliary draft model for semantic alignment with those leveraging N-gram patterns to capture repetition, unlocking distinct potential for reasoning under test-time scaling. However, current hybrid strategies remain heuristic and coarse, failing to effectively harness the semantic richness of draft models and the reusability of N-gram patterns. We hope these findings can provide practical guidance for the design of speculative decoding strategies in test-time scaling, particularly in balancing adaptability, efficiency, and robustness for complex reasoning tasks.

\section{Related Work}
\label{sec:related}

Our research is positioned at the intersection of two rapidly evolving domains in LLMs: test-time scaling for enhanced reasoning and speculative decoding for inference acceleration. This section reviews the key developments in both areas and highlights the gap our work aims to fill.

\subsection{Test-Time Scaling: Enhancing Reasoning via Computation}

The paradigm of test-time scaling~\citep{DBLP:conf/iclr/snelllee25}, or inference-time search, enhances model performance by allocating more computational resources at inference time. This principle can be realized through two main avenues: (1) specialized training that endows a model with an inherent capability to dynamically scale its computation, or (2) external, training-free frameworks that guide a fixed model~\citep{DBLP:arxiv/zhanglyu25}.

The former approach, inherent test-time scaling, is embodied by the development of ``slow thinking" or System 2 reasoning models. Models such as DeepSeek-R1~\citep{deepseekr1}, Qwen3~\citep{DBLP:arxiv/yangli25}, and the OpenAI o1 series~\citep{openai-o1} are explicitly trained to internalize multi-step reasoning. This is often achieved via reinforcement learning on complex tasks, teaching the model to endogenously allocate more computation before producing an output.

Complementing this inherent capacity are external test-time scaling frameworks. These techniques apply an outer loop of control to a base model, further amplifying its reasoning performance without modifying its parameters. These external methods can be applied to any model, including those already possessing inherent slow-thinking capabilities. 
This paper focuses on analyzing two foundational categories of such frameworks:
\begin{itemize}
    [leftmargin=*]
    \item \textbf{Sampling-Based Methods:} The most prominent approach in this category is Best-of-N sampling~\citep{DBLP:conf/iclr/snelllee25}, where a model generates $N$ candidate outputs, and a verifier or the model itself selects the most plausible one. A more sophisticated variant, Self-Consistency~\citep{DBLP:conf/iclr/wangwei23}, generates multiple reasoning paths and selects the final answer by a majority vote, demonstrating significant improvements in arithmetic and commonsense reasoning. These methods operate on the principle that a model is more likely to generate a correct reasoning path than an incorrect one, and sampling multiple paths increases the probability of finding a valid solution.

    \item \textbf{Iterative Refinement:} This class of methods involves a multi-step process where the model progressively improves its own output. For instance, Self-Refine~\citep{DBLP:conf/nips/madaantandon23} enables a model to generate an initial response, critique it, and then use that feedback to produce a refined answer in a subsequent pass. A more streamlined approach is multi-round thinking~\citep{DBLP:arxiv/tianzhao25}, which simplifies the feedback loop. Instead of generating an explicit critique, it provides only the final answer from the previous turn as context, prompting the model to re-evaluate its conclusion. This method is designed to break ``cognitive inertia" by discarding the prior reasoning chain, forcing the model to approach the problem anew.
    
\end{itemize}

Other search algorithms, such as Tree-of-Thoughts (ToT)~\citep{DBLP:journals/nips/yaoyu23}, Graph-of-Thoughts~\citep{DBLP:conf/aaai/BestaBKGPGGLNNH24}, and those based on Monte-Carlo Tree Search (MCTS)~\citep{lin2025}, can be understood as sophisticated hybrids of these foundational principles. For instance, ToT explores multiple candidate reasoning steps (a form of sampling) and evaluates them to construct a solution tree, while MCTS iteratively builds and refines a search policy. By combining multi-candidate generation with progressive construction and evaluation, they represent a powerful but more complex class of techniques built upon the core ideas of sampling and refinement.


\subsection{Efficient Reasoning}
While proven effective for unlocking the advanced reasoning capabilities of LLMs, test-time scaling paradigms are inherently resource-intensive. This significant increase in computational overhead and inference latency presents a substantial challenge to their practical deployment in real-world, time-sensitive applications. Efficient reasoning offers a promising path to alleviate this burden by generating smarter, more concise reasoning with less computation \citep{DBLP:arxiv/WuYao25,DBLP:arxiv/SuiChuang25,DBLP:icml_workshop/XingLi}. Research in this area can be organized into three main categories: (i) Model-based efficient reasoning \citep{DBLP:arxiv/WangShen25,DBLP:arxiv/LouSun25}, which aims to optimize full-length reasoning models into more concise variants or directly train compact models specialized for fast and effective reasoning. (ii) Reasoning output-based efficient reasoning \citep{DBLP:arxiv/HaoSukhbaatar24, DBLP:arxiv/PanDai25,DBLP:arxiv/LinLi25}, which dynamically reduces the computational work during inference by adaptively shortening the reasoning chain and eliminating unnecessary steps. (iii) Input prompts-based efficient reasoning \citep{DBLP:arxiv/MaHe25,DBLP:arxiv/LiangZhong25}, which enhances efficiency by tailoring the reasoning process according to the properties of the input, such as its specified length constraints.

However, the speculative decoding methods for accelerating test-time scaling examined in this paper is orthogonal to above efficient reasoning methodologies. Fundamentally, speculative decoding represents a lossless acceleration technique that preserves the original model's output distribution exactly, ensuring that the quality and characteristics of generated reasoning remain unchanged. Unlike efficient reasoning approaches that achieve speedup by modifying the reasoning process itself, whether through model quantization, output compression, or prompt engineering, speculative decoding operates at the inference level without altering the underlying reasoning methodology or sacrificing any aspect of the original model's performance.

\subsection{Speculative Decoding: Accelerating Autoregressive Generation}

Speculative decoding has emerged as a state-of-the-art (SOTA) solution for reducing LLM inference latency without altering the model's output distribution. The core idea involves using a smaller, faster draft model or mechanism to generate a sequence of candidate tokens, which are then verified in parallel by the large, original target model in a single forward pass. This allows for multiple tokens to be generated per target model evaluation, leading to significant speedups. The landscape of speculative decoding is diverse and can be classified into three main families:

\begin{itemize}
    [leftmargin=*]
    \item \textbf{Model-Based Speculative Sampling:} This is the classic approach where a smaller, general-purpose language model (e.g., a distilled or simply smaller version of the target model) serves as the drafter~\citep{DBLP:conf/icml/leviathankalman23,DBLP:arxiv/chenborgeaud23}. While straightforward to implement, its efficiency depends on the alignment between the draft and target models' probability distributions, which may not be optimal for specialized tasks.

    \item \textbf{Training-Based Speculative Decoding:} To improve alignment, this family of methods involves training a specialized draft model. Approaches like Medusa~\citep{DBLP:arxiv/caili24} achieve this by adding multiple, non-autoregressive decoding ``heads" directly to the target model's architecture. In contrast, the EAGLE series of models~\citep{DBLP:conf/icml/liwei24a, DBLP:conf/emnlp/liwei24b, DBLP:arxiv/liwei25} trains a separate, lightweight autoregressive drafter using knowledge distillation. This drafter learns to approximate the target model's next-token distribution by processing its internal hidden states, enabling high accuracy without altering the original model's structure. While these methods can achieve very high acceptance rates, they require a significant upfront investment in training resources and data.

    \item \textbf {N-gram-based Methods:} This lightweight and adaptive category of methods is distinguished by being entirely training-free. Instead of using a learned model, it builds a "draft" by retrieving token sequences from a dynamic cache of recent outputs. The simplest implementations use a dictionary or lookup table of recently generated n-grams to predict future tokens~\citep{DBLP:conf/naacl/hezhong24,DBLP:conf/kdd/zhaoxie24}. More advanced techniques enhance retrieval efficiency by employing sophisticated data structures, such as the suffix automaton~\citep{DBLP:conf/acl/huwang25} or the suffix tree~\citep{DBLP:arxiv/oliarojia24}. By its very nature, this approach is particularly effective at capturing and accelerating the repetitive patterns and local regularities common in generated text.
\end{itemize}

While these methods have been benchmarked on generic tasks like text summarization and single-turn question answering, their interplay with test-time scaling frameworks remains largely unexplored. The highly structured and often repetitive nature of reasoning chains generated by methods like Best-of-N~\citep{DBLP:conf/iclr/snelllee25} and multi-round thinking~\citep{DBLP:arxiv/tianzhao25} presents a unique environment where the trade-offs between these different speculative decoding families are unknown. Our work directly addresses this gap by systematically evaluating their performance in these latency-critical, redundancy-rich reasoning scenarios.

\begin{figure}[h]
    \centering
    \includegraphics[width=1.0\linewidth]{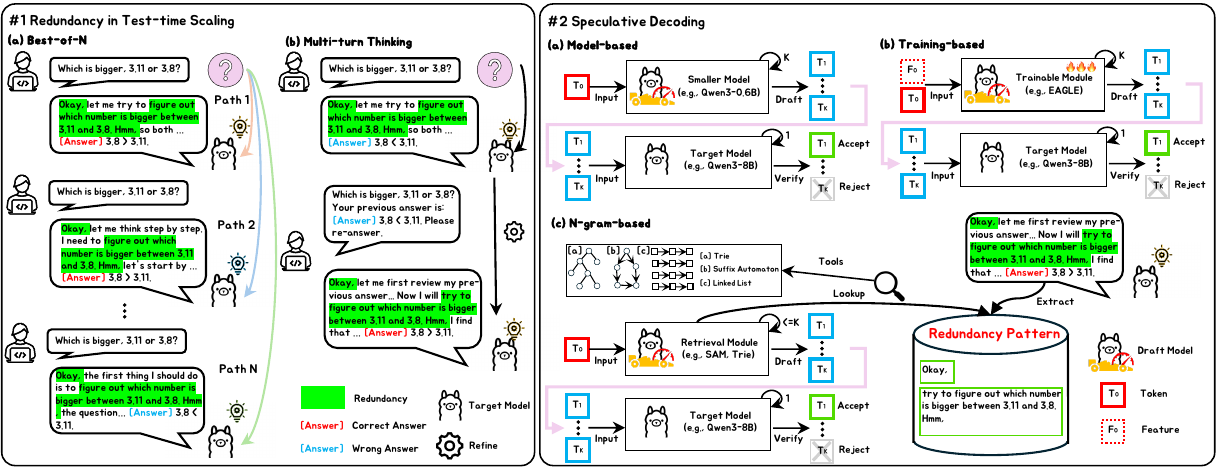}
    \caption{The framework of speculative decoding mehods for efficient LLM test-time scaling.  (\#1) An overview of test-time scaling frameworks (Best-of-N and multi-turn thinking) and their inherent redundancies. (\#2) Illustrations of three categories of speculative decoding methods. \textit{Model-based method} takes tokens as input and use a smaller, homologous model as the draft model to generate speculative token sequences. \textit{Training-based method} utilizes both input tokens and hidden-layer features, employing an additionally trained lightweight draft model to predict future tokens. \textit{N-gram-based method} leverages tools such as suffix automata to retrieve repeated patterns in history for efficient draft generation.}
    \label{ssy0801:overall_framework}
\end{figure}

\section{Benchmark Construction}
\label{sec:background}

To empirically evaluate the efficacy of speculative decoding in accelerating slow-thinking processes, we first establish a clear and robust benchmark. This section details the three core components of our experimental setup, including the test-time scaling frameworks we aim to accelerate, the datasets used for evaluation, and the speculative decoding algorithms under comparison. An overall framework is illustrated in \Figref{ssy0801:overall_framework}.

\subsection{Benchmarked Test-time Scaling Frameworks}

As discussed in \Secref{sec:related}, several frameworks have been proposed to enhance LLM performance by allocating more computation at inference time. This work selects two particularly representative and widely-adopted frameworks, namely Best-of-N and multi-round thinking, as the basis of our benchmark. Their significant computational cost, which is the very source of their improved reasoning ability, makes them ideal testbeds for evaluating the potential of acceleration techniques. This section provides a detailed mechanical overview of each framework.

\subsubsection{Best-of-N}
The Best-of-N~\citep{DBLP:arxiv/zhanglyu25} is a foundational technique designed to improve output quality by leveraging a generate-and-rank mechanism. Instead of relying on a single, potentially flawed output from the model, Best-of-N operates in two distinct stages:

\begin{itemize}
    [leftmargin=*]
    \item \textbf{Generation:} Given a request, the LLM is prompted to generate $N$ independent candidate solutions or reasoning paths. This is typically achieved by sampling from the model's output distribution multiple times with a non-zero temperature to ensure diversity. The collection of these $N$ samples represents a broad exploration of the potential solution space.

    \item \textbf{Selection/Verification:} After generating the $N$ candidates, a verifier module is used to score and select the best solution among them. The verifier can range from a simple, heuristic-based checker to a more sophisticated, separately trained reward model. In its most canonical form, a process-based verifier might assess the correctness of each intermediate step in a solution, rather than just evaluating the final answer. The candidate with the highest score is then chosen as the final output.
\end{itemize}

\subsubsection{Multi-round Thinking}

Multi-round thinking~\citep{DBLP:arxiv/tianzhao25} improves reasoning through iterative refinement. This method is inspired by the human cognitive process of reconsideration and self-correction. It enhances a model's reasoning by prompting it to reflect on and improve its own previous answers over multiple rounds, as follows:

\begin{itemize}
    [leftmargin=*]
    \item \textbf{Initial Response (Round 1):} The LLM generates an initial answer to the user's prompt. This represents the model's initial or intuitive response.

    \item \textbf{Iterative Refinement (Rounds 2 to \textit{M}):} The process is repeated for a pre-defined total of $M$ rounds. In each subsequent round (from 2 to $M$), the model is prompted to reflect on and refine its previous output. The new prompt is constructed by augmenting the original question with the answer from the preceding round. For example, the prompt for the second round takes the following structure:
    \begin{quote}
        \small 
        \ttfamily 
        \{Original Question\} \\[0.5ex]
        The assistant's previous answer is: \{Round 1 Answer\} \\[0.5ex]
        Please re-answer.
    \end{quote}
    This mechanism forces the model to reconsider its initial reasoning path, allowing it to self-correct errors and elaborate on its response with each iteration.
\end{itemize}

\subsection{Benchmarked Datasets}
\label{ssy0801:info_dataset}
Having defined the computational frameworks that expand reasoning effort at inference time, we now turn to the datasets used for evaluation. The choice of datasets is critical to ensure that our benchmark measures reasoning capabilities across a diverse set of challenging problems. To this end, we have selected four widely-recognized benchmarks that have become staples in recent literature for evaluating the limits of LLM reasoning capabilities. Our evaluation suite is composed of 120 samples extracted from the following datasets:
\begin{itemize}
    [leftmargin=*]
    \item \textbf{AIME 2024 and AIME 2025:} The American Invitational Mathematics Examination (AIME) \citep{aime} features challenging competition-level mathematics problems that demand creative and rigorous problem-solving skills. We use the official problems from the 2024 and 2025 competitions, utilizing the full set of 30 problems from each.

    \item \textbf{MATH-500:} A subset of the MATH dataset~\citep{math500}, a comprehensive benchmark of mathematics problems ranging from pre-algebra to calculus and beyond. It is the standard for evaluating mathematical reasoning in LLMs. Given the considerable size of the full 500-problem dataset, we selected the first 30 problems to ensure a focused and manageable evaluation.

    \item \textbf{GPQA:} The GPQA dataset~\citep{gpqa} consists of graduate-level, Google-proof Q\&A problems across biology, physics, and chemistry, designed to test internalized knowledge and complex reasoning. The full dataset contains 448 expert-level questions, representing a substantial and challenging collection. To maintain consistency with our other evaluation subsets, we selected the first 30 problems from this dataset.
\end{itemize}

This benchmark serves as a challenging testbed for our core experiments. Specifically, we will leverage these 120 problems to evaluate the acceleration effects of speculative decoding methods when applied to the computationally intensive \textbf{Best-of-N} and \textbf{multi-round thinking} frameworks. This allows us to quantify the efficiency gains in practical, high-stakes reasoning scenarios.

\begin{table*}[htbp]
    \centering
    \caption{Summary of different speculative decoding methods. The table outlines each method's key idea, category (\textit{model-based}, \textit{training-based}, or \textit{N-gram-based}), speculation structure (\textit{Linear} or \textit{Tree}), ability to reuse repetitive content during generation, and its compatibility with \textit{Greedy} decoding and speculative \textit{Sampling}. }
    \label{ssy0801:speculative_methods_summary}
    \resizebox{1.0\textwidth}{!}{%
        \begin{tabular}{l p{6cm} c c c c c}
            \toprule
            \multicolumn{1}{l}{\multirow{2}{*}{\textbf{Methods}}} 
            & \multirow{2}{*}{\textbf{Key Idea}} 
            & \multicolumn{2}{c}{\textbf{Drafting}} 
            & \multicolumn{2}{c}{\textbf{Verification}} 
            & \multirow{2}{*}{\textbf{Category}} \\
            \cmidrule(lr){3-4} \cmidrule(lr){5-6}
            & 
            & Speculation & Reuse 
            & Greedy & Sampling \\

            \midrule

            \makecell[l]{SpS~\citep{DBLP:arxiv/chenborgeaud23}}& 
            Speculative decoding with smaller homologous models and rejection sampling. & 
            Linear & 
            \redno &
            \greenyes & 
            \greenyes & 
             
            Model-based\\
            \addlinespace[1.5pt]
            
            \makecell[l]{EAGLE-3~\citep{DBLP:arxiv/liwei25}} & 
            Introducing trainable lightweight auto-regressive heads on intermediate layers. & 
            Tree & 
            \redno &
            \greenyes & 
            \greenyes & 
             
            Training-based\\
            \addlinespace[1.5pt]
            
            \makecell[l]{PLD~\citep{saxena2023prompt}} & 
            Generates drafts by reusing N-gram patterns extracted from the input prompt. & 
            Linear & 
            \redno &
            \greenyes & 
            \redno & 
             
            N-gram-based\\
            \addlinespace[1.5pt]
            
            \makecell[l]{REST~\cite{DBLP:conf/naacl/hezhong24}} & 
            Drafting based on an external datastore for retrieving likely next tokens. & 
            Tree & 
            \redno &
            \greenyes & 
            \greenyes & 
             
            N-gram-based\\
            \addlinespace[1.5pt]
            
            \makecell[l]{Lookahead~\citep{DBLP:arxiv/fubailis24}} & 
            Generating multiple n-gram candidates (extracted from generation) in parallel. & 
            Tree & 
            \greenyes & 
            \greenyes & 
            \redno & 
            
            N-gram-based\\
            \addlinespace[1.5pt]
            
            \makecell[l]{PIA~\citep{DBLP:conf/kdd/zhaoxie24}} & 
            Speculative decoding via Trie-structured context caching. & 
            Tree & 
            \greenyes &
            \greenyes & 
            \greenyes & 
             
            N-gram-based\\
            \addlinespace[1.5pt]
            
            \makecell[l]{SAM~\citep{DBLP:conf/acl/huwang25}} & 
            Generating drafts using a suffix automaton derived from the generation history. & 
            Linear & 
            \greenyes &
            \greenyes & 
            \greenyes & 
             
            N-gram-based\\
            \addlinespace[1.5pt]
            
            \makecell[l]{Recycling~\citep{DBLP:conf/acl/luowang25}} & 
            Generating drafts by recycling top-K next tokens from history. & 
            Tree & 
            \redno &
            \greenyes & 
            \greenyes & 
             
            N-gram-based\\
            
            \bottomrule
        \end{tabular}%
    }
\end{table*}

\subsection{Benchmarked Speculative Decoding Algorithms}
\label{ssy0801:diff_methods}
With the test-time scaling frameworks and evaluation datasets established, the final component of our benchmark is the set of speculative decoding algorithms we will evaluate. We benchmark a diverse set of methods, which can be broadly categorized into three groups, including model-based speculative sampling (e.g., SpS), training-based speculative decoding (e.g., EAGLE-3), and N-gram-based methods (e.g., PLD, REST, Lookahead, PIA, SAM, and Recycling). This collection of methods encompasses almost all existing speculative decoding techniques compatible with reasoning models, capturing the full spectrum of current SOTA solutions. We provide a summary of the characteristics of above methods in Table \ref{ssy0801:speculative_methods_summary}. Detailed descriptions of each method are presented below.

\begin{itemize}[leftmargin=*]
\item \textbf{Model-based Method}
    \begin{itemize}[label=\textbullet, leftmargin=*, itemsep=0pt, parsep=2pt]
        \item \textit{Speculative Sampling (SpS)}~\citep{DBLP:arxiv/chenborgeaud23}: As a pioneering and representative speculative decoding method, SpS employs a smaller, faster draft model to generate a sequence of candidate tokens. The larger, more powerful target model then validates these tokens in parallel using a rejection sampling mechanism. This process guarantees that the output distribution remains identical to that of the target model alone, ensuring lossless generation.
    \end{itemize}
\end{itemize}

\begin{itemize}[leftmargin=*]
\item \textbf{Training-based Method}
    \begin{itemize}[label=$\bullet$, leftmargin=*, itemsep=0pt, parsep=2pt]
    \item \textit{EAGLE-3}~\citep{DBLP:arxiv/liwei25}: This is an advanced speculative decoding approach that dispenses with a separate draft model. Instead, it attaches lightweight auto-regressive heads to the target model's intermediate layers to generate drafts. This design minimizes extra memory and storage overhead while achieving significant speedups by enabling parallel token verification.
    \end{itemize}
\end{itemize}

\begin{itemize}[leftmargin=*]
\item \textbf{N-gram-based Method}
    \begin{itemize}[label=\textbullet,leftmargin=*, itemsep=0pt, parsep=2pt]
    \item \textit{Prompt Lookup Decoding (PLD)}~\citep{saxena2023prompt}: This method enhances generation speed by leveraging the input prompt itself as a source for draft tokens. PLD identifies n-grams from the prompt that match the sequence generated so far and uses them as speculative candidates. PLD is particularly effective in tasks where the output is likely to contain substrings from the input, such as in summarization or question-answering.
    
    \item \textit{Retrieval-Based Speculative Decoding (REST)}~\citep{DBLP:conf/naacl/hezhong24}: REST utilizes a datastore of example sequences and a retrieval model to find continuations that are relevant to the current generation context. These retrieved continuations serve as the speculative draft, which is then verified by the target model. This approach allows the model to draft long, coherent sequences when a similar context is found in the datastore.
    
    \item \textit{Lookahead}~\citep{DBLP:arxiv/fubailis24}: A parallel decoding algorithm that generates and validates multiple n-grams simultaneously. Unlike traditional speculative decoding which drafts a single sequence, Lookahead maintains multiple candidate continuations. It uses a tree-based structure to efficiently verify these candidates in a single forward pass of the target model, allowing it to accelerate generation without an explicit draft model.
    
    \item \textit{PIA}~\citep{DBLP:conf/kdd/zhaoxie24}: This method utilizes a trie structure to manage and retrieve contextual information from the prompt and decoding context efficiently. It supports tree-based speculative decoding, where multiple candidate token sequences are generated and verified in parallel, thereby accelerating the inference process by leveraging recurring patterns in the input context.

    \item \textit{SAM}~\citep{DBLP:conf/acl/huwang25}: SAM employs a suffix automaton to maintain the contextual history of the generated sequence. This data structure allows for the effective identification of previously generated token sequences (suffixes) that match the current context, enabling linear speculative decoding by reusing these matched suffixes as draft candidates.

    \item \textit{Recycling}~\citep{DBLP:conf/acl/luowang25}: Recycling enhances speculative decoding by caching the top-K most probable subsequent tokens for each token verified by the target model. This cache, constructed from the highest logits observed during past generation steps, serves as a high-quality source of draft candidates for future occurrences of the same token, thereby improving the acceptance rate.
    \end{itemize}
\end{itemize}

\begin{itemize}[leftmargin=*]
\item \textbf{Hybrid Method}
    \begin{itemize}[label=\textbullet, leftmargin=*, itemsep=0pt, parsep=2pt]
    \item \textit{SAM with EAGLE-3 (SAM[EAGLE-3])}~\citep{DBLP:conf/acl/huwang25}: This approach is a hybrid strategy that dynamically switches between the SAM and EAGLE-3 frameworks. When SAM identifies a matched suffix that is too short to be beneficial for speculation, the system defaults to EAGLE-3, which uses a small auxiliary draft model to generate speculative tokens. This adaptability optimizes performance across diverse contexts.
    \end{itemize}
\end{itemize}

\section{Speculative Decoding for Reasoning under Test-Time Scaling}

\subsection{Experiment Setup}
We investigate the effect of speculative decoding methods in reasoning models DeepSeek-R1-Distill-Llama-8B (DSL-8B) \citep{deepseekr1} and Qwen3-8B (QW3-8B) \citep{DBLP:arxiv/yangli25}, employing the widely-used test-time scaling paradigms of multi-round thinking (with 2 rounds) and BoN sampling (with 4 trajectories). We evaluate various speculative decoding methods on the popular reasoning datasets AIME24, AIME25 \citep{aime}, MATH500 \citep{math500}, and GPQA \citep{gpqa}. More details about datasets can be found in Section \ref{ssy0801:info_dataset}. The evaluation pipeline from prior work \citep{DBLP:conf/acl/xiayang24} is adopted to ensure fair comparison and reproducibility. For the reasoning models, we use the float16 data type and apply greedy/stochastic decoding with a batch size of 1, following the original settings of the evaluated speculative decoding methods. The evaluation metrics include: \textit{Mean Accepted Tokens (MAT)}, which measures the average number of tokens accepted per speculative decoding step, and the \textit{Walltime Speedup Ratio (Speed)}, which quantifies the inference efficiency gain relative to vanilla autoregressive decoding. The evaluated methods are categorized into three groups (details on the categories are provided in Section \ref{ssy0801:diff_methods}), including model-based speculative sampling (e.g., SpS), training-based speculative decoding (e.g., EAGLE-3), and N-gram-based methods (e.g., PLD, REST, Lookahead, PIA, SAM, and Recycling). We followed the default parameters as specified in their original implementations. All experiments are conducted on a server equipped with a 16-core Intel Xeon Silver 4309Y CPU and a single NVIDIA RTX A6000 GPU (48GB).

\begin{table}[h]
  \centering
  \caption{Performance comparison of speculative decoding methods for reasoning models under multi-round thinking framework with different temperature $T$ (\colorbox{red!10}{\textbf{Best}}, \colorbox{green!20}{\underline{Second Best}}). }
  \resizebox{\textwidth}{!}{
    \begin{tabular}{c|c|cccccccc|cc}
    \toprule[1.5pt]
    \multirow{2}[4]{*}{Model} & \multicolumn{1}{c|}{Bench} & \multicolumn{2}{c}{AIME24} & \multicolumn{2}{c}{AIME25} & \multicolumn{2}{c}{MATH500} & \multicolumn{2}{c|}{GPQA} & \multicolumn{2}{c}{Overall} \\
    \cmidrule{2-12}          & Method      & MAT   & Speed & MAT   & Speed & MAT   & Speed & MAT   & Speed & MAT   & Speed \\
    \midrule
    \multicolumn{1}{c|}{\multirow{9}{*}{\makecell{DSL-8B\\($T=0$)}}}
          & \multicolumn{1}{c|}{AR} & 1.00     & 1.00$\times$     & 1.00     & 1.00$\times$     & 1.00     & 1.00$\times$     & 1.00     & 1.00$\times$     & 1.00     & 1.00$\times$ \\
          & EAGLE-3 & 2.26  & 1.56$\times$  & 2.24  & 1.53$\times$  & 2.63  &\cellcolor{green!20}\underline{2.86$\times$}  & 2.46  & 1.61$\times$  & 2.35  & 1.93$\times$ \\
          & SAM${\text{[EAGLE-3]}}$ & \cellcolor{red!10}\textbf{3.91}  & \cellcolor{red!10}\textbf{3.09$\times$}  & \cellcolor{red!10}\textbf{4.50}   & \cellcolor{red!10}\textbf{3.85}$\times$  & \cellcolor{red!10}\textbf{4.94}  & \cellcolor{red!10}\textbf{4.11$\times$}  & \cellcolor{red!10}\textbf{7.00}    & \cellcolor{red!10}\textbf{4.70$\times$}   & \cellcolor{red!10}\textbf{4.72}  & \cellcolor{red!10}\textbf{3.97$\times$} \\
          & SAM   &2.64  & \cellcolor{green!20}\underline{2.41$\times$}  &\cellcolor{green!20}\underline{3.05}  &\cellcolor{green!20}\underline{2.96$\times$}  & 2.60   & 2.14$\times$  &\cellcolor{green!20}\underline{3.57}  &\cellcolor{green!20}\underline{3.20$\times$}   & 2.93  &\cellcolor{green!20}\underline{2.66$\times$} \\
          & Recycling &\cellcolor{green!20}\underline{2.98}  & 2.08$\times$  & 2.98  & 2.05$\times$  &\cellcolor{green!20}\underline{2.97}  &2.18$\times$  & 3.02  & 2.08$\times$  &\cellcolor{green!20}\underline{2.99}  & 2.10$\times$ \\
          & PLD   & 2.14  & 1.76$\times$  & 2.38  & 1.89$\times$  & 2.25  & 1.71$\times$  & 2.65  & 2.02$\times$  & 2.33  & 1.84$\times$ \\
          & REST  &1.33	&1.00$\times$	&1.35	&0.95$\times$	&1.37	&1.08$\times$	&1.31	&1.00$\times$	&1.34	&1.01$\times$ \\
          & Lookahead & 2.21  & 1.59$\times$  & 2.26  & 1.53$\times$  & 2.28  & 1.55$\times$  & 2.44  & 1.62$\times$  & 2.28  & 1.57$\times$ \\
          & PIA   & 1.84  & 1.51$\times$  & 1.96  & 1.63$\times$  & 1.91  & 1.48$\times$  & 2.15  & 1.63$\times$  & 1.95  & 1.56$\times$ \\
    \midrule
    \multicolumn{1}{c|}{\multirow{7}{*}{\makecell{DSL-8B\\($T=0.6$)}}}
          & \multicolumn{1}{c|}{AR} & 1.00     & 1.00$\times$     & 1.00     & 1.00$\times$     & 1.00     & 1.00$\times$     & 1.00     & 1.00$\times$     & 1.00     & 1.00$\times$ \\
          & EAGLE-3  &2.21   &1.31$\times$   &2.17   &1.50$\times$   &\cellcolor{green!20}\underline{3.21}   &\cellcolor{green!20}\underline{3.02$\times$}   &2.45   &1.61$\times$   &2.33   &1.91$\times$ \\
          & SAM${\text{[EAGLE-3]}}$   &\cellcolor{green!20}\underline{2.54}   &\cellcolor{green!20}\underline{1.82$\times$}   &\cellcolor{green!20}\underline{2.48}   &\cellcolor{green!20}\underline{1.89$\times$}   &\cellcolor{red!10}\textbf{3.28}   &\cellcolor{red!10}\textbf{3.13$\times$}   &\cellcolor{red!10}\textbf{2.86}   &\cellcolor{red!10}\textbf{2.15$\times$}   &\cellcolor{green!20}\underline{2.66}   &\cellcolor{red!10}\textbf{2.29$\times$} \\
          & SAM     &1.84   &1.65$\times$   &1.84   &1.65$\times$	  &1.93   &1.71$\times$   &1.93   &1.73$\times$	  &1.87   &1.69$\times$\\
          & Recycling   &\cellcolor{red!10}\textbf{2.84}   &\cellcolor{red!10}\textbf{1.91$\times$}	  &\cellcolor{red!10}\textbf{2.83}   &\cellcolor{red!10}\textbf{1.93$\times$}	  &2.88   &2.03$\times$	  &\cellcolor{green!20}\underline{2.81}   &\cellcolor{green!20}\underline{1.98$\times$}	  &\cellcolor{red!10}\textbf{2.84}   &\cellcolor{green!20}\underline{1.96$\times$}\\
          & REST   &1.35 & 1.00$\times$ & 1.37 & 0.97$\times$ & 1.38 & 1.03$\times$ & 1.35 & 1.00$\times$ & 1.36 & 1.00$\times$\\
          & PIA     &1.58   &1.26$\times$	  &1.58   &1.27$\times$   &1.63   &1.39$\times$	  &1.68   &1.36$\times$	  &1.61   &1.33$\times$\\
    \midrule
    \multirow{10}[0]{*}{\makecell{QW3-8B\\($T=0$)}} 
    & \multicolumn{1}{c|}{AR} & 1.00     & 1.00$\times$     & 1.00     & 1.00$\times$     & 1.00     & 1.00$\times$     & 1.00     & 1.00$\times$     & 1.00     & 1.00$\times$ \\
    & EAGLE-3 & 4.31  & \cellcolor{green!20}\underline{2.80$\times$}   & 4.37  &\cellcolor{green!20}\underline{2.87$\times$}  & 4.50   &\cellcolor{green!20}\underline{3.04$\times$}  & 4.39  & 2.91$\times$  & 4.38  &\cellcolor{green!20}\underline{2.91$\times$} \\
          & SAM${\text{[EAGLE-3]}}$ & \cellcolor{green!20}\underline{4.40}   & \cellcolor{red!10}\textbf{3.08$\times$} &\cellcolor{green!20}\underline{4.47}  & \cellcolor{red!10}\textbf{3.18$\times$}  &\cellcolor{green!20}\underline{4.82}  &\cellcolor{red!10}\textbf{3.47$\times$}  &\cellcolor{green!20}\underline{5.90}   &\cellcolor{red!10}\textbf{4.17$\times$ } &\cellcolor{green!20}\underline{4.76}  &\cellcolor{red!10}\textbf{3.49$\times$} \\
          & SAM   & 2.15  & 1.97$\times$  & 2.27  & 2.07$\times$  & 2.18  & 1.95$\times$  & 3.19  &\cellcolor{green!20}\underline{3.11$\times$}  & 2.37  & 2.28$\times$ \\
          & Recycling & 3.02  & 2.08$\times$  & 2.98  & 2.13$\times$  & 3.00     & 2.20$\times$   & 3.07  & 2.17$\times$  & 3.01  & 2.15$\times$ \\
          & PLD   & 1.95  & 1.61$\times$  & 1.97  & 1.66$\times$  & 1.91  & 1.63$\times$  & 2.49  & 2.04$\times$  & 2.05  & 1.74$\times$ \\
          & SpS   & \cellcolor{red!10}\textbf{7.40}   & 0.93$\times$  & \cellcolor{red!10}\textbf{7.50}   & 0.96$\times$  & \cellcolor{red!10}\textbf{6.69 } & 0.77$\times$  & \cellcolor{red!10}\textbf{6.33}  & 0.83$\times$  & \cellcolor{red!10}\textbf{7.07}  & 0.87$\times$ \\
          & REST  & 1.39	& 1.12$\times$	& 1.39	& 1.14$\times$	& 1.42	& 1.16$\times$	& 1.33	& 1.09$\times$	& 1.38	& 1.13$\times$ \\
          & Lookahead & 2.14  & 1.54$\times$  & 2.12  & 1.54$\times$  & 2.10   & 1.58$\times$  & 2.28  & 1.69$\times$  & 2.15  & 1.59$\times$ \\
          & PIA   & 1.83  & 1.48$\times$  & 1.96  & 1.68$\times$  & 1.89  & 1.63$\times$  & 2.08  & 1.72$\times$  & 1.93  & 1.63$\times$ \\
    \midrule
    \multirow{8}[0]{*}{\makecell{QW3-8B\\($T=0.6$)}}
    & \multicolumn{1}{c|}{AR} & 1.00     & 1.00$\times$     & 1.00     & 1.00$\times$     & 1.00     & 1.00$\times$     & 1.00     & 1.00$\times$     & 1.00     & 1.00$\times$ \\
    & EAGLE-3    &\cellcolor{green!20}\underline{4.11}   &\cellcolor{green!20}\underline{2.61$\times$}	&\cellcolor{green!20}\underline{4.21} &\cellcolor{green!20}\underline{2.71$\times$}	&\cellcolor{green!20}\underline{4.32} &\cellcolor{green!20}\underline{2.87$\times$}	&\cellcolor{green!20}\underline{4.03} &\cellcolor{green!20}\underline{2.68$\times$}	&\cellcolor{green!20}\underline{4.16} &\cellcolor{green!20}\textbf{2.73$\times$}\\
          & SAM${\text{[EAGLE-3]}}$ &3.92 &\cellcolor{red!10}\textbf{2.68}$\times$	&3.93 &\cellcolor{red!10}\textbf{2.72$\times$}	&4.19 &\cellcolor{red!10}\textbf{2.93$\times$}	&3.98 &\cellcolor{red!10}\textbf{2.82$\times$}	&3.97 &\cellcolor{red!10}\textbf{2.79$\times$} \\
          & SAM   &1.91 &1.73$\times$	&1.95 &1.78$\times$	&1.93 &1.74$\times$	&2.09 &1.86$\times$	&1.96 &1.78$\times$\\
          & Recycling &2.88 &2.00$\times$	&2.92 &2.04$\times$	&2.90 &2.11$\times$	&2.92 &2.08$\times$	&2.90 &2.06$\times$\\
          & SpS   &\cellcolor{red!10}\textbf{6.22} & 0.91$\times$ &\cellcolor{red!10}\textbf{6.54} & 0.95$\times$ &\cellcolor{red!10}\textbf{6.17} & 0.80$\times$ &\cellcolor{red!10}\textbf{6.32} & 0.86$\times$ &\cellcolor{red!10}\textbf{6.34} & 0.88$\times$\\
          & REST  &1.41 &1.08$\times$	&1.41 &1.10$\times$	&1.42 &1.12$\times$	&1.36 &1.03$\times$	&1.40 &1.08$\times$\\
          & PIA   &1.69 &1.40$\times$	&1.70 &1.45$\times$	&1.71 &1.45$\times$	&1.75 &1.48$\times$	&1.71 &1.44$\times$\\
          \bottomrule[1.5pt]
    \end{tabular}}
  \label{ssy0801:thinktwice}%
\end{table}

\subsection{Overall Comparison}
In this section, we evaluate speculative decoding methods for reasoning under test-time scaling paradigms. The performance of different methods under multi-round thinking is presented in Table \ref{ssy0801:thinktwice}. As DSL-8B does not have a smaller model variant, results for the combination of SpS and DSL-8B are not provided. For QW3-8B, we evaluate SpS using Qwen3-0.6B (QW3-0.6B) \citep{DBLP:arxiv/yangli25} as the draft model. Additionally, under the speculative sampling setting, methods such as Lookahead do not support non-greedy decoding, hence their non-greedy variants are not included in Table \ref{ssy0801:thinktwice}. Table \ref{ssy0801:bon_T_06} illustrates the performance of speculative decoding methods under the BoN paradigm. Since the BoN paradigm necessitates temperature sampling to generate diverse reasoning trajectories, greedy decoding is not evaluated in this context.

\textbf{$\bullet$ The adaptability of the training-based method is constrained by its training process.} For the training-based method, we evaluate the SOTA EAGLE-3, which supports reasoning model acceleration. Other training-based methods, such as EAGLE-1/2 \citep{DBLP:conf/icml/liwei24a,DBLP:conf/emnlp/liwei24b} and Medusa \citep{DBLP:arxiv/caili24}, are not included in our comparison because they do not provide corresponding draft models for reasoning models. For DSL-8B, we used a draft model trained by the EAGLE-3 team\footnote{\url{https://huggingface.co/yuhuili/EAGLE3-DeepSeek-R1-Distill-LLaMA-8B}}, while for QW3-8B, we use a draft model provided by the Tengyunw team\footnote{\url{https://huggingface.co/Tengyunw/qwen3_8b_eagle3}}. As shown in Table \ref{ssy0801:thinktwice} and Table \ref{ssy0801:bon_T_06}, EAGLE-3 achieves considerable speedup across different scenarios. For example, in the multi-round thinking paradigm with greedy decoding, EAGLE-3 achieves a speedup of up to 3.04$\times$ when accelerating QW3-8B. However, we also observe some performance deviations from expectations. Specifically, the average speedup ratio $(T=0)$ in Table \ref{ssy0801:thinktwice} decreases by 34\% when running EAGLE-3 on DSL-8B compared to QW3-8B (2.91$\times$$\rightarrow$1.93$\times$). Moreover, the overall MAT $(T=0)$ of EAGLE-3 on DSL-8B in Table \ref{ssy0801:thinktwice} is approximately half of that on QW3-8B (4.38$\rightarrow$2.35). Similar trends are observed in the temperature sampling setting and Table \ref{ssy0801:bon_T_06}. These differences exceed the typical performance variation that can be attributed to intrinsic model factors such as differences in vocabulary or architectural design. We hypothesize two potential reasons for this phenomenon: (i) As noted by the Tengyunw team, they used 600k samples to train the draft model for QW3-8B, which is larger than the dataset used by the EAGLE-3 team for training the DSL-8B draft model. A \textit{larger training set can improve the alignment} between the draft and target models' output distributions, leading to better performance. (ii) The original EAGLE-3 evaluation focused on short-generation scenarios. We conjecture that the training process for the draft model may \textit{not adequately model long-generation settings} (for example, larger position indices may receive insufficient training updates), leading to performance degradation in long-generation scenarios of the reasoning model.

\textbf{$\bullet$ Training-based method is insensitive to temperature.} As the temperature $T$ increases, the target model's output distribution becomes more uniform and diverse, making it more challenging for the draft model to accurately predict the subsequent tokens compared to greedy decoding. Despite this increased difficulty, the training-based method EAGLE-3 demonstrates considerable robustness to temperature variations. For instance, under the multi-round thinking paradigm, when the $T$ is raised from 0 to 0.6, the average speedup of EAGLE-3 on QW3-8B decreases by only 6\% (2.91$\times$$\rightarrow$2.73$\times$). This relative stability suggests that the training-based method has the potential to generalize across different levels of decoding stochasticity.

\begin{takeaway}
While the training-based methods demonstrate promising acceleration potential, their performance is inherently tied to the training process, raising practical concerns about their adaptability to diverse reasoning scenarios.
\end{takeaway}

\begin{table}[htbp]
  \centering
  \caption{Performance comparison of speculative decoding methods for reasoning models under BoN framework with different temperature $T$ (\colorbox{red!10}{\textbf{Best}}, \colorbox{green!20}{\underline{Second Best}}).}
  \resizebox{\textwidth}{!}{
    \begin{tabular}{c|c|cccccccc|cc}
    \toprule[1.5pt]
    \multirow{2}[4]{*}{Model} & \multicolumn{1}{c|}{Bench} & \multicolumn{2}{c}{AIME24} & \multicolumn{2}{c}{AIME25} & \multicolumn{2}{c}{MATH500} & \multicolumn{2}{c|}{GPQA} & \multicolumn{2}{c}{Overall} \\
\cmidrule{2-12}          & Method      & MAT   & Speed & MAT   & Speed & MAT   & Speed & MAT   & Speed & MAT   & Speed \\
    \midrule
    \multicolumn{1}{c|}{\multirow{7}{*}{\makecell{DSL-8B\\$(T=0.6)$}}}
    & \multicolumn{1}{c|}{AR} & 1.00     & 1.00$\times$     & 1.00     & 1.00$\times$     & 1.00     & 1.00$\times$     & 1.00     & 1.00$\times$     & 1.00     & 1.00$\times$ \\
          & EAGLE-3   & 2.19 & 1.17$\times$ & 2.18 & 1.43$\times$ & \cellcolor{green!20}\underline{3.09} & \cellcolor{green!20}\underline{2.71$\times$} & 2.37 & 1.52$\times$ & 2.31 & 1.74$\times$\\
          & SAM${\text{[EAGLE-3]}}$   & \cellcolor{green!20}\underline{2.44} & 1.53$\times$ & \cellcolor{green!20}\underline{2.41} & \cellcolor{green!20}\underline{1.72$\times$} &\cellcolor{red!10}\textbf{3.31} & \cellcolor{red!10}\textbf{3.05$\times$} & \cellcolor{green!20}\underline{2.72} & \cellcolor{red!10}\textbf{1.96$\times$} & \cellcolor{green!20}\underline{2.56} & \cellcolor{red!10}\textbf{2.10$\times$} \\
          & SAM   & 1.89 & \cellcolor{green!20}\underline{1.62$\times$} & 1.86 & 1.62$\times$ & 1.97 & 1.77$\times$ & 2.00 & 1.75$\times$ & 1.91 & 1.70$\times$  \\
          & Recycling  & \cellcolor{red!10}\textbf{2.81} & \cellcolor{red!10}\textbf{1.91$\times$} & \cellcolor{red!10}\textbf{2.80} & \cellcolor{red!10}\textbf{1.86$\times$} & 2.88 & 2.04$\times$ & \cellcolor{red!10}\textbf{2.80} & \cellcolor{green!20}\underline{1.93$\times$} & \cellcolor{red!10}\textbf{2.82} & \cellcolor{green!20}\underline{1.94$\times$} \\
          & REST    & 1.36 & 0.99$\times$ & 1.38 & 0.99$\times$ & 1.40 & 1.07$\times$ & 1.33 & 0.98$\times$ & 1.37 & 1.01$\times$\\
          & PIA     & 1.53 & 1.18$\times$ & 1.55 & 1.21$\times$ & 1.62 & 1.35$\times$ & 1.68 & 1.37$\times$ & 1.57 & 1.20$\times$\\
    \midrule
    \multirow{8}[0]{*}{\makecell{QW3-8B\\$(T=0.6)$}} 
    & \multicolumn{1}{c|}{AR} & 1.00     & 1.00$\times$     & 1.00     & 1.00$\times$     & 1.00     & 1.00$\times$     & 1.00     & 1.00$\times$     & 1.00     & 1.00$\times$ \\
    & EAGLE-3  & \cellcolor{green!20}\underline{4.09} & \cellcolor{green!20}\underline{2.63$\times$} & \cellcolor{green!20}\underline{4.17} & \cellcolor{green!20}\underline{2.64$\times$} & \cellcolor{green!20}\underline{4.30} & \cellcolor{green!20}\underline{2.85$\times$} & \cellcolor{green!20}\underline{4.00} & \cellcolor{green!20}\underline{2.64$\times$} & \cellcolor{green!20}\underline{4.12} & \cellcolor{green!20}\underline{2.70$\times$}  \\
          & SAM${\text{[EAGLE-3]}}$ & 3.81 & \cellcolor{red!10}\textbf{2.66$\times$} & 3.91 & \cellcolor{red!10}\textbf{2.72$\times$} & 4.18 & \cellcolor{red!10}\textbf{2.99$\times$} & 3.87 & \cellcolor{red!10}\textbf{2.68$\times$} & 3.91 & \cellcolor{red!10}\textbf{2.77$\times$} \\
          & SAM  & 2.05 & 1.88$\times$ & 2.04 & 1.88$\times$ & 2.17 & 2.01$\times$ & 2.19 & 2.00$\times$ & 2.10 & 1.95$\times$ \\
          & Recycling & 2.90 & 2.03$\times$ & 2.88 & 1.99$\times$ & 2.90 & 2.06$\times$ & 2.91 & 2.04$\times$ & 2.89 & 2.03$\times$ \\
          & SpS   &\cellcolor{red!10}\textbf{6.05} & 0.93$\times$ & \cellcolor{red!10}\textbf{6.22} & 0.95$\times$ & \cellcolor{red!10}\textbf{6.01} & 0.80$\times$ & \cellcolor{red!10}\textbf{5.87} & 0.86$\times$ & \cellcolor{red!10}\textbf{6.05} & 0.88$\times$\\
          & REST  & 1.41 & 1.12$\times$ & 1.40 & 1.10$\times$ & 1.42 & 1.11$\times$ & 1.36 & 1.07$\times$ & 1.40 & 1.10$\times$ \\
          & PIA   & 1.67 & 1.38$\times$ & 1.70 & 1.41$\times$ & 1.72 & 1.44$\times$ & 1.76 & 1.45$\times$ & 1.70 & 1.42$\times$ \\
          \bottomrule[1.5pt]
    \end{tabular}}
  \label{ssy0801:bon_T_06}%
\end{table}

\subsubsection{N-gram-based Method --- Token N-gram}
\textbf{$\bullet$ Among token N-gram methods, which identify repeated patterns at the token level, SAM most efficiently captures redundant patterns.} For token N-gram methods, we assess SAM, PLD, REST, Lookahead, and PIA. Overall, SAM achieves the best performance among these token N-gram methods, even rivaling the training-based method EAGLE-3. For instance, under the greedy decoding setting with multi-round thinking, SAM achieves a 38\% higher overall speedup ratio than EAGLE-3 on DSL-8B (2.66$\times$$\rightarrow$1.93$\times$). This advantage primarily stems from the high efficiency of SAM in suffix matching, as well as the repetitive patterns inherent in reasoning model outputs. Other methods, such as PLD, PIA, and Lookahead, also demonstrate considerable potential, achieving around 1.5$\times$ speedup across various test-time scaling paradigms. However, due to the higher time complexity of their N-gram matching mechanisms, they still fall short of SAM in acceleration. REST performs relatively poorly in our experiments. We hypothesize this is because the external chat datastore used by REST has limited overlap with the long cot reasoning dataset, resulting in fewer effective token reuse opportunities.

\textbf{$\bullet$ Sampling temperature significantly affects token N-gram-based methods.} We note that token N-gram methods exhibit notable sensitivity to sampling temperature. Taking SAM as an example, under the multi-round thinking paradigm, when the temperature increases from 0 to 0.6, its overall speedup on QW3-8B drops by approximately 22\% (from 2.28$\times$ to 1.78$\times$), and on MAT it decreases from 2.37 to 1.96, representing a reduction of about 17\%. This highlights a key trade-off: while these methods are particularly effective at leveraging the repetitive patterns in reasoning model outputs under test-time scaling, their acceleration benefits tend to decrease as output diversity increases.

\begin{takeaway}
Token N-gram methods excel at capturing emergent redundancy in reasoning trajectories under test-time scaling paradigms, yet this advantage is accompanied by sensitivity to sampling temperature.
\end{takeaway}

\subsubsection{N-gram-based Method --- Probabilistic N-gram}
\textbf{$\bullet$ Probabilistic N-gram method Recycling, which captures the co-occurrence of top-K next tokens through logits, achieves the highest MAT among N-gram-based methods.} As shown in Table \ref{ssy0801:thinktwice}, Recycling achieves performance comparable to the strongest retrieval-based method SAM. Under greedy decoding, Recycling slightly lags behind SAM in speedup ratio, yet attains a higher MAT. We attribute this to Recycling's higher computational complexity in draft generation. Specifically, Recycling employs a tree-structured speculative verification, resulting in significantly higher computational latency compared to SAM's linear verification. Moreover, Recycling speculates 81 tokens, which is nearly twice the number of tokens speculated by SAM (40 tokens), which further contributes to its overall computational overhead. A similar phenomenon is observed in SpS: the use of a source-consistent draft model that shares the same origin as the target model ensures high output distribution alignment, resulting in substantially higher MAT. However, the draft model's relatively large size (i.e., 0.6B) introduces a computational bottleneck that limits speedup gains.

\textbf{$\bullet$ Probabilistic N-gram method  exhibits low sensitivity to temperature sampling.} As the temperature increases from 0 to 0.6 in the multi-round thinking paradigm, Recycling's overall speedup ratio on QW3-8B decreases only from 2.15 to 2.06—a decline of less than 5\%. In contrast, SAM's speedup drops by 22\% under the same temperature shift. We hypothesize that this robustness stems from Recycling's preservation of approximate next-token prediction logits, specifically the top-k vocabulary distribution, which enables substantial overlap between its tree-based speculative tokens and the target model's sampled tokens even as temperature varies. Overall, unlike token N-gram methods that rely on historical token sequences, Recycling innovatively leverages historical logits, endowing it with exceptional promise for reasoning tasks where temperature sampling is prevalent.

\begin{takeaway}
The probabilistic N-gram method offers superior stability and higher draft token acceptance rates among N-gram-based methods, yet its computational bottleneck prevents acceptance gains from translating into proportional speedup.
\end{takeaway}

\subsubsection{Hybrid Speculative Decoding: Integrating Training-based and N-gram Method}
\textbf{$\bullet$ Hybrid speculative decoding can leverage the strengths of different methods.} For the hybrid speculative decoding, we conduct evaluations on SAM${\text{[EAGLE-3]}}$, which integrates methods EAGLE-3 and SAM (detailed hybrid strategies are provided in Section \ref{ssy0801:diff_methods}). 
As shown in Table \ref{ssy0801:thinktwice}, SAM${\text{[EAGLE-3]}}$ achieves the highest overall speedup ratio across all scenarios. Under the greedy decoding setting, the hybrid method SAM${\text{[EAGLE-3]}}$ demonstrates significant improvements over individual methods. Within the multi-turn thinking paradigm, SAM${\text{[EAGLE-3]}}$ delivers speedup gains of 20\% and 53\% compared to EAGLE-3 and SAM on QW3-8B, respectively. The reason behind this improvement is straightforward: SAM${\text{[EAGLE-3]}}$ preserves EAGLE-3's powerful token speculation capabilities while addressing its limitation in capturing repetitive patterns that emerge during dynamic generation.

\textbf{$\bullet$ Hybrid speculative decoding retains the limitations of its components.} While inheriting the strengths of SAM, SAM${\text{[EAGLE-3]}}$ also retains its key drawback, namely, a high sensitivity to sampling temperature. When the sampling temperature increases from 0 to 0.6, the speedup ratio of SAM${\text{[EAGLE-3]}}$ under the multi-round thinking paradigm decreases by 42\% on DSL-8B (3.97$\times$$\rightarrow$2.29$\times$) and by 20\% on QW3-8B (3.49$\times$$\rightarrow$2.79$\times$), respectively. Furthermore, as observed in Table \ref{ssy0801:thinktwice} and Table \ref{ssy0801:bon_T_06}, the performance gain of SAM${\text{[EAGLE-3]}}$ over EAGLE-3 is less significant. This indicates that under temperature sampling with $T=0.6$, the accuracy of using SAM for speculative decoding is not satisfactory, leading to the overall performance of SAM${\text{[EAGLE-3]}}$ being largely dependent on that of EAGLE-3. Despite certain limitations, the consistent SOTA performance of SAM${\text{[EAGLE-3]}}$ across various evaluation scenarios underscores its practical effectiveness. By integrating the strengths of different methods, the hybrid method achieves well-balanced performance across various settings, highlighting its potential for reasoning with test-time scaling.

\begin{takeaway}
Hybrid speculative decoding uniquely combines the semantic alignment strengths of training-based methods with the repetitive pattern capture capabilities of N-gram methods, unlocking distinct potential for reasoning with test-time scaling.
\end{takeaway}

\subsubsection{Model-based Method}
\textbf{$\bullet$ Model-based method SpS generates high-quality drafts, but faces computational bottlenecks.} As shown in Table \ref{ssy0801:thinktwice} and Table \ref{ssy0801:bon_T_06}, SpS achieves the highest MAT. For example, in the multi-round thinking scenario with QW3-8B ($T=0$), SpS's overall MAT is 1.49 times that of the second-best method, and this substantial advantage remains consistent across other settings. This superiority primarily stems from SpS employing a smaller, homogenous (architecturally aligned) model as the draft model, which ensures a high degree of alignment between the output distributions of the draft and target models, thereby significantly improving token acceptance rates. However, this high MAT does not translate into proportional speedup. The primary bottleneck lies in the computational cost of generating drafts. Despite being smaller, such as a 0.6B model, the draft model is still too heavy compared to the target models like 8B or 14B, particularly under tight latency constraints. As a result, the time saved by reducing target model evaluations is largely offset by the overhead of running the draft model, which limits the overall acceleration.

\begin{takeaway}
Model-based method SpS achieves high MAT by leveraging a well-aligned draft model, but its acceleration is limited due to the relatively small size difference between the draft model (e.g., 0.6B) and the large target model (e.g., 8B).
\end{takeaway}

\section{Further Analysis} \label{sec:analysis}
In this section, we further analyze speculative decoding in terms of acceleration across reasoning turns, computational overhead of key modules, the potential of hybrid strategies, and performance on models of different scales, offering insights into its practical efficiency and limitations.

\begin{figure}[h]
    \centering
    \subcaptionbox{\label{ssy0801:speed_turns_multi_think_00}}{\includegraphics[width=0.47\linewidth]{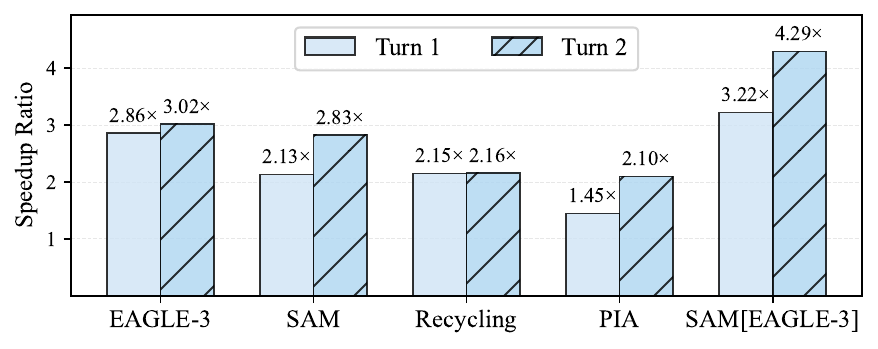}\vspace{-5pt}}
    \hspace{-2pt}
    \subcaptionbox{\label{ssy0801:speed_turns_multi_think_06}}{\includegraphics[width=0.47\linewidth]{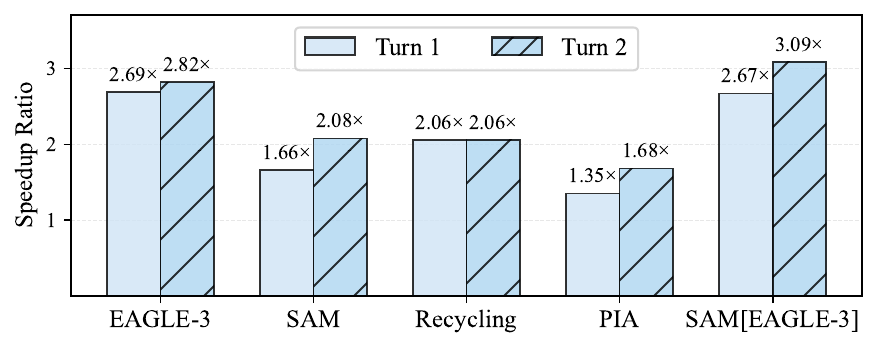}\vspace{-5pt}}
    \subcaptionbox{\label{ssy0801:speed_turns_BoN_06}}{\includegraphics[width=0.95\linewidth]{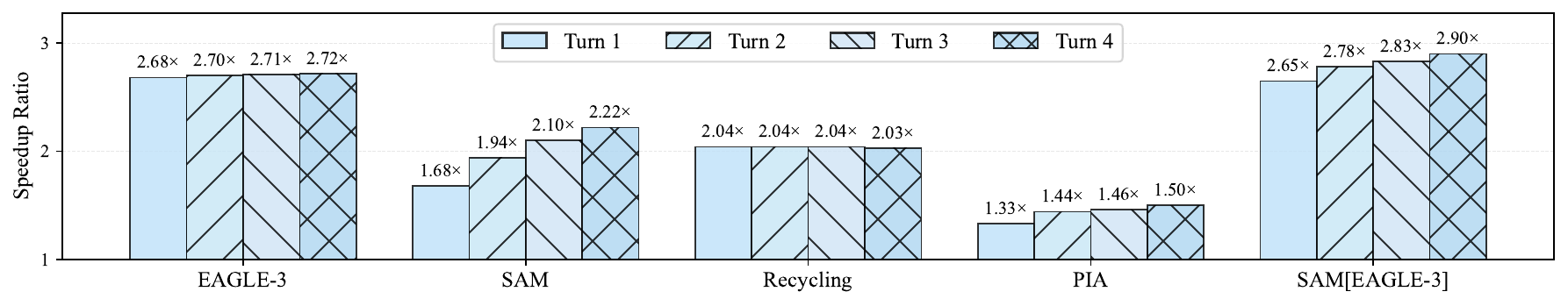}\vspace{-5pt}}
    \vspace{-5pt}
    \caption{Speedup ratio across different turns under different settings across various temperature $T$. (a) Multi-round thinking ($T=0$). (b) Multi-round thinking ($T=0.6$). (c) BoN ($T=0.6$).}
    \label{ssy0801:speed_turns}
\end{figure}

\noindent\textbf{$\bullet$ Comparison of acceleration across turns}. \Figref{ssy0801:speed_turns} compares the acceleration ratios of different methods across multiple reasoning turns. A key observation is that retrieval-based methods exhibit a steadily increasing acceleration ratio as the number of turns grows. As illustrated in \Figref{ssy0801:speed_turns_multi_think_00}, SAM and PIA show acceleration ratio increases of 33\% (2.13$\times$$\rightarrow$2.83$\times$) and 45\% (1.45$\times$$\rightarrow$2.10$\times$) in the second turn over the first, respectively. This trend demonstrates their ability to effectively identify and exploit repetitive patterns that naturally emerge during test-time scaling. Specifically, by retrieving and reusing relevant intermediate results or reasoning traces from prior turns, these methods reduce the number of required decoding steps in subsequent turns, leading to progressively greater efficiency gains over time. In contrast, non-retrieval-based methods exhibit nearly constant acceleration ratios across turns, indicating that each turn is processed independently without leveraging reusable patterns from prior decoding steps. They fail to exploit the correlations between reasoning turns, resulting in no meaningful efficiency improvement as the number of turns increases.

\begin{takeaway}
Retrieval-based methods PIA and SAM achieve progressive acceleration by reusing past computations, enabling efficiency that scales across turns.
\end{takeaway}

\begin{figure}[h]
    \centering
    \subcaptionbox{\label{ssy0801:compare_three_highlight_00}}{\includegraphics[width=0.47\linewidth]{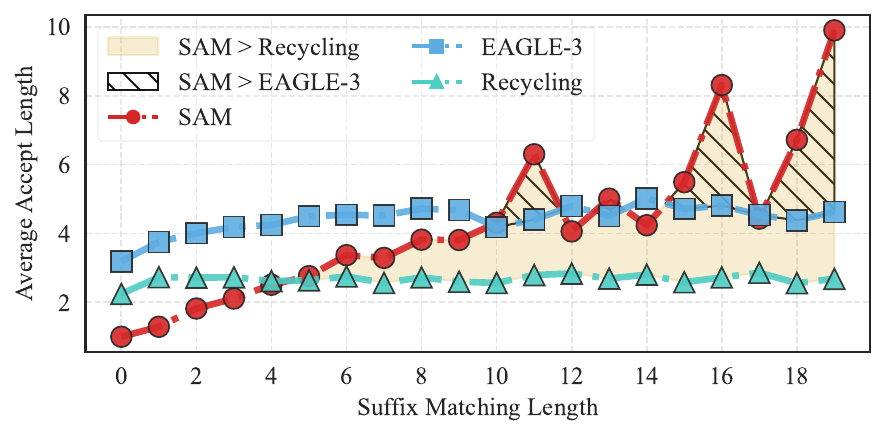}\vspace{-5pt}}
    \hspace{-2pt}
    \subcaptionbox{\label{ssy0801:compare_three_highlight_06}}{\includegraphics[width=0.47\linewidth]{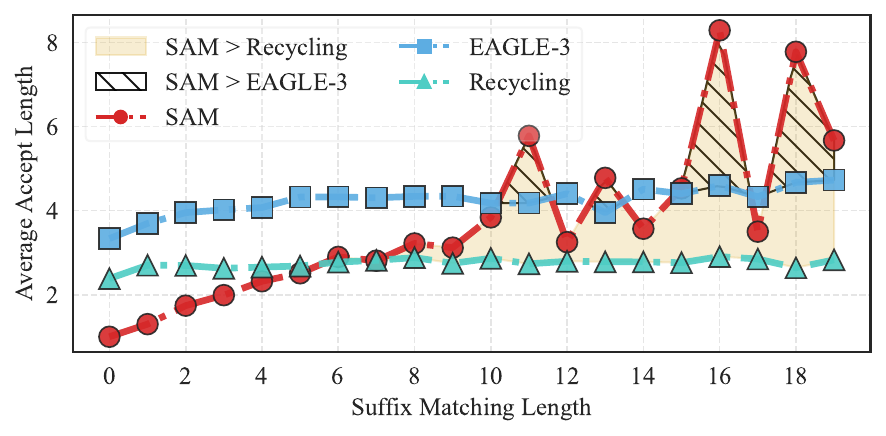}\vspace{-5pt}}
    \caption{Average accept length under different suffix matching lengths across various temperature $T$. (a) $T=0$. (b) $T=0.6$.}
    \label{ssy0801:accept_length_matching}
\end{figure}

\noindent\textbf{$\bullet$ Potential analysis of hybrid speculative decoding}. We present the average accept length across different suffix matching lengths and temperature settings in \Figref{ssy0801:accept_length_matching}. The results indicate that SAM demonstrates strong potential when the matching length exceeds 10. For instance, under greedy decoding ($T=0$), SAM achieves an average accept length greater than 9 at a suffix matching length of 11, which is nearly twice the average accept length of EAGLE-3. However, we also observe notable fluctuations in SAM’s average accept length as the matching suffix length increases. This sensitivity suggests that the performance of speculative decoding can vary significantly depending on the hybrid strategy (based on the matched suffix length). It is promising to design more refined hybrid strategies to unlock SAM's potential for long-sequence matching in test-time scaling, which will further improve the performance of the hybrid speculative decoding.

\begin{figure}
    \raggedright
    \subcaptionbox{\label{ssy0801:time_modules_8b}}{
    \includegraphics[width=0.47\linewidth]{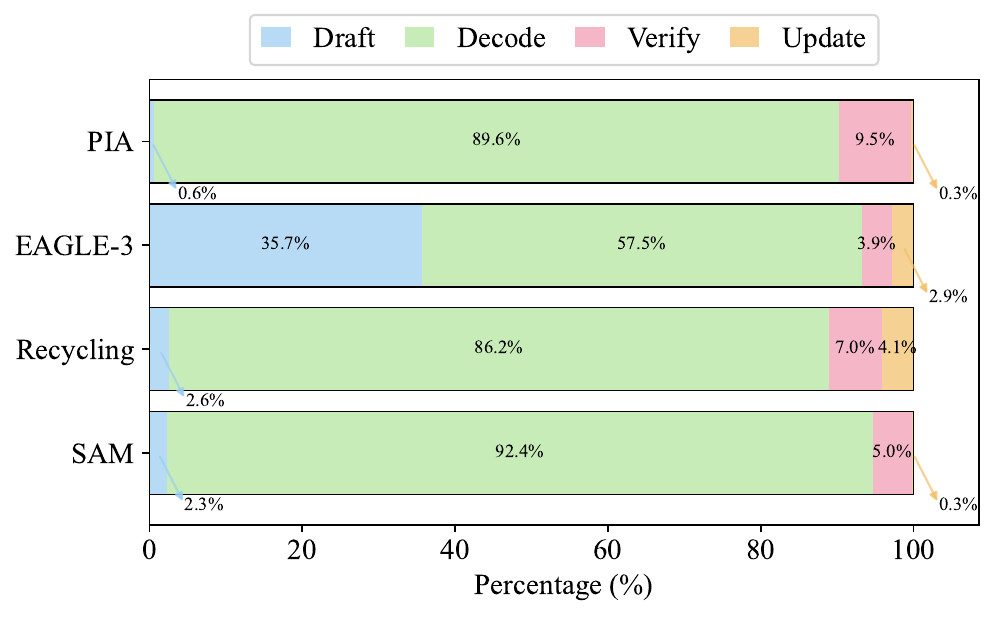}\vspace{-5pt}}\hspace{1em}
    \subcaptionbox{\label{ssy0801:time_modules_14b}}{
    \includegraphics[width=0.47\linewidth]{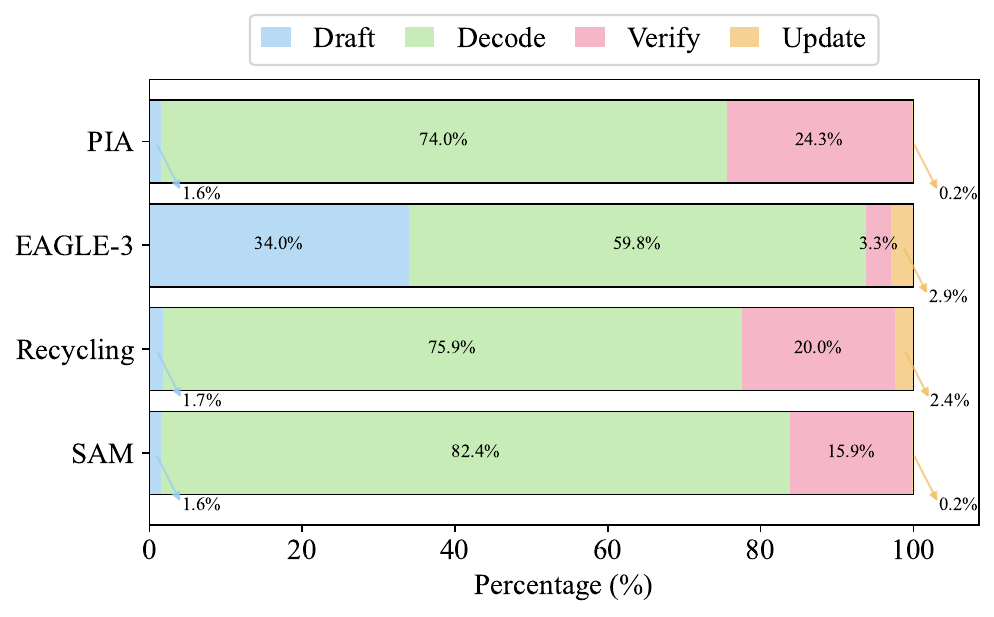}\vspace{-5pt}}
    \caption{Analysis of time consumption proportions across modules in speculative decoding with different model sizes. (a) QW3-8B. (b) QW3-14B.}
    \label{ssy0801:time_modules}
\end{figure}

\begin{table}[h]
  \centering
  \caption{Performance comparison of speculative decoding methods on reasoning models with different scales across various temperature $T$ (\colorbox{red!10}{\textbf{Best}}, \colorbox{green!20}{\underline{Second Best}}).}
  \resizebox{\textwidth}{!}{
    \begin{tabular}{c|c|cccccccc|cc}
    \toprule[1.5pt]
    \multirow{2}[4]{*}{Model} & \multicolumn{1}{c|}{Bench} & \multicolumn{2}{c}{AIME24} & \multicolumn{2}{c}{AIME25} & \multicolumn{2}{c}{MATH500} & \multicolumn{2}{c|}{GPQA} & \multicolumn{2}{c}{Overall} \\
\cmidrule{2-12}          & Method      & MAT   & Speed & MAT   & Speed & MAT   & Speed & MAT   & Speed & MAT   & Speed \\
    \midrule
    \multicolumn{12}{c}{\textit{Under the Multi-turn Thinking Paradigm}} \\
    \midrule
    \multicolumn{1}{c|}{\multirow{9}{*}{\makecell{QW3-4B\\($T=0.6$)}}}
    & \multicolumn{1}{c|}{AR} & 1.00     & 1.00$\times$     & 1.00     & 1.00$\times$     & 1.00     & 1.00$\times$     & 1.00     & 1.00$\times$     & 1.00     & 1.00$\times$ \\
    & EAGLE-3 & \cellcolor{red!10}\textbf{3.45} & \cellcolor{green!20}\underline{2.28$\times$} & \cellcolor{red!10}\textbf{3.48} & \cellcolor{green!20}\underline{2.22$\times$} & \cellcolor{red!10}\textbf{3.54} & \cellcolor{green!20}\underline{2.38$\times$} & \cellcolor{green!20}\underline{3.31} & 2.20$\times$ & \cellcolor{red!10}\textbf{3.44} & \cellcolor{green!20}\underline{2.27$\times$} \\
          & SAM${\text{[EAGLE-3]}}$ &  \cellcolor{green!20}\underline{3.40} & \cellcolor{red!10}\textbf{2.37$\times$} & \cellcolor{green!20}\underline{3.40} & \cellcolor{red!10}\textbf{2.28$\times$} & \cellcolor{green!20}\underline{3.50} & \cellcolor{red!10}\textbf{2.55$\times$} & \cellcolor{red!10}\textbf{3.47} & \cellcolor{red!10}\textbf{2.39$\times$} & \cellcolor{green!20}\underline{3.43} & \cellcolor{red!10}\textbf{2.40$\times$}\\
          & SAM   & 1.99 & 1.92$\times$ & 1.94 & 1.81$\times$ & 1.95 & 1.82$\times$ & 2.19 & 1.99$\times$ & 2.01 & 1.89$\times$\\
          & Recycling & 2.92 & 2.20$\times$ & 2.91 & 2.11$\times$ & 2.92 & 2.38$\times$ & 2.94 & \cellcolor{green!20}\underline{2.25$\times$} & 2.92 & 2.24$\times$ \\
          & REST  & 1.41 & 1.12$\times$ & 1.42 & 1.10$\times$ & 1.44 & 1.15$\times$ & 1.35 & 1.04$\times$ & 1.40 & 1.10$\times$ \\
          & PIA   &  1.71 & 1.45$\times$ & 1.70 & 1.39$\times$ & 1.69 & 1.42$\times$ & 1.78 & 1.46$\times$ & 1.72 & 1.43$\times$\\
    \midrule
    \multicolumn{1}{c|}{\multirow{7}{*}{\makecell{QW3-14B\\($T=0.6$)}}}
    & \multicolumn{1}{c|}{AR} & 1.00     & 1.00$\times$     & 1.00     & 1.00$\times$     & 1.00     & 1.00$\times$     & 1.00     & 1.00$\times$     & 1.00     & 1.00$\times$ \\
          & EAGLE-3  & \cellcolor{green!20}\underline{3.28} & \cellcolor{red!10}\textbf{2.26$\times$} & \cellcolor{green!20}\underline{3.38} & \cellcolor{red!10}\textbf{2.31$\times$} & \cellcolor{green!20}\underline{3.36} & \cellcolor{green!20}\underline{2.27$\times$} & 3.05 & 2.10$\times$ & \cellcolor{green!20}\underline{3.28} & \cellcolor{green!20}\underline{2.23$\times$}\\
          & SAM${\text{[EAGLE-3]}}$   &  3.18 & \cellcolor{green!20}\underline{2.24$\times$} & 3.26 & \cellcolor{green!20}\underline{2.29$\times$} & 3.31 & \cellcolor{red!10}\textbf{2.31$\times$} & \cellcolor{green!20}\underline{3.12} & \cellcolor{red!10}\textbf{2.15$\times$} & 3.21 & \cellcolor{red!10}\textbf{2.25$\times$}\\
          & SAM     & 1.84 & 1.70$\times$ & 1.86 & 1.72$\times$ & 1.85 & 1.66$\times$ & 1.90 & 1.73$\times$ & 1.86 & 1.70$\times$\\
          & Recycling   & 2.86 & 2.10$\times$ & 2.86 & 2.08$\times$ & 2.89 & 2.17$\times$ & 2.85 & \cellcolor{green!20}\underline{2.13$\times$} & 2.86 & 2.12$\times$\\
          & REST   & 1.41 & 1.17$\times$ & 1.40 & 1.17$\times$ & 1.42 & 1.17$\times$ & 1.33 & 1.11$\times$ & 1.39 & 1.15$\times$\\
          & SpS & \cellcolor{red!10}\textbf{5.78} & 1.22$\times$ & \cellcolor{red!10}\textbf{5.96} & 1.25$\times$ & \cellcolor{red!10}\textbf{5.73} & 1.11$\times$ & \cellcolor{red!10}\textbf{4.59} & 1.02$\times$ & \cellcolor{red!10}\textbf{5.59} & 1.14$\times$\\
          & PIA     & 1.62 & 1.43$\times$ & 1.69 & 1.44$\times$ & 1.66 & 1.45$\times$ & 1.69 & 1.46$\times$ & 1.66 & 1.45$\times$\\
    \midrule
    \multicolumn{12}{c}{\textit{Under the BoN Paradigm}} \\
    \midrule
    \multirow{10}[0]{*}{\makecell{QW3-4B\\($T=0.6$)}}
    & \multicolumn{1}{c|}{AR} & 1.00     & 1.00$\times$     & 1.00     & 1.00$\times$     & 1.00     & 1.00$\times$     & 1.00     & 1.00$\times$     & 1.00     & 1.00$\times$ \\
    & EAGLE-3 & \cellcolor{red!10}\textbf{3.45} & \cellcolor{green!20}\underline{2.24$\times$} & \cellcolor{red!10}\textbf{3.48} & \cellcolor{green!20}\underline{2.23$\times$} & \cellcolor{green!20}\underline{3.52} & \cellcolor{green!20}\underline{2.31$\times$} & \cellcolor{green!20}\underline{3.33} & 2.11$\times$ & \cellcolor{red!10}\textbf{3.44} & \cellcolor{green!20}\underline{2.22$\times$}\\
          & SAM${\text{[EAGLE-3]}}$ & \cellcolor{green!20}\underline{3.37} & \cellcolor{red!10}\textbf{2.37$\times$} & \cellcolor{green!20}\underline{3.37} & \cellcolor{red!10}\textbf{2.33$\times$} & \cellcolor{red!10}\textbf{3.58} & \cellcolor{red!10}\textbf{2.61$\times$} & \cellcolor{red!10}\textbf{3.38} & \cellcolor{red!10}\textbf{2.36$\times$} & \cellcolor{green!20}\underline{3.40} & \cellcolor{red!10}\textbf{2.42$\times$}\\
          & SAM   & 2.12 & 1.94$\times$ & 2.13 & 1.92$\times$ & 2.22 & 2.07$\times$ & 2.32 & 2.07$\times$ & 2.18 & 2.00$\times$\\
          & Recycling & 2.92 & 2.14$\times$ & 2.91 & 2.11$\times$ & 2.92 & 2.31$\times$ & 2.96 & \cellcolor{green!20}\underline{2.20$\times$} & 2.93 & 2.19$\times$\\
          & REST  & 1.42 & 1.11$\times$ & 1.41 & 1.11$\times$ & 1.44 & 1.16$\times$ & 1.38 & 1.03$\times$ & 1.41 & 1.10$\times$\\
          & PIA   & 1.70 & 1.40$\times$ & 1.72 & 1.41$\times$ & 1.75 & 1.50$\times$ & 1.83 & 1.51$\times$ & 1.74 & 1.46$\times$\\
    \midrule
    \multirow{8}[0]{*}{\makecell{QW3-14B\\($T=0.6$)}} 
    & \multicolumn{1}{c|}{AR} & 1.00     & 1.00$\times$     & 1.00     & 1.00$\times$     & 1.00     & 1.00$\times$     & 1.00     & 1.00$\times$     & 1.00     & 1.00$\times$ \\
    & EAGLE-3    & \cellcolor{green!20}\underline{3.28} & \cellcolor{green!20}\underline{2.27$\times$} & \cellcolor{green!20}\underline{3.31} & \cellcolor{green!20}\underline{2.27$\times$} & 3.33 & \cellcolor{green!20}\underline{2.27$\times$} & 3.04 & 2.07$\times$ & \cellcolor{green!20}\underline{3.24} & \cellcolor{green!20}\underline{2.22$\times$}\\
          & SAM${\text{[EAGLE-3]}}$ &3.21 & \cellcolor{red!10}\textbf{2.33$\times$} & 3.22 & \cellcolor{red!10}\textbf{2.34$\times$} & \cellcolor{green!20}\underline{3.35} & \cellcolor{red!10}\textbf{2.47$\times$} & \cellcolor{green!20}\underline{3.10} & \cellcolor{red!10}\textbf{2.20$\times$} & 3.21 & \cellcolor{red!10}\textbf{2.33$\times$}\\
          & SAM   &1.98 & 1.84$\times$ & 1.99 & 1.87$\times$ & 2.09 & 1.93$\times$ & 2.05 & 1.85$\times$ & 2.01 & 1.88$\times$\\
          & Recycling &2.86 & 2.12$\times$ & 2.86 &  2.10$\times$ & 2.89 & 2.20$\times$ & 2.87 & \cellcolor{green!20}\underline{2.15$\times$} & 2.87 & 2.15$\times$\\
          & SpS   &\cellcolor{red!10}\textbf{5.76} & 1.22$\times$ & \cellcolor{red!10}\textbf{5.76} & 1.24$\times$ & \cellcolor{red!10}\textbf{5.62} & 1.09$\times$ & \cellcolor{red!10}\textbf{4.87} & 1.04$\times$ & \cellcolor{red!10}\textbf{5.53} & 1.14$\times$\\
          & REST  &1.40 & 1.16$\times$ & 1.40 & 1.16$\times$ & 1.42 & 1.18$\times$ & 1.34 & 1.11$\times$ & 1.39 & 1.15$\times$\\
          & PIA   &1.64 & 1.42$\times$ & 1.69 & 1.44$\times$ & 1.70 & 1.46$\times$ & 1.73 & 1.45$\times$ & 1.68 & 1.45$\times$\\
          \bottomrule[1.5pt]
    \end{tabular}}
  \label{ssy0801:different_size}%
\end{table}

\begin{takeaway}
SAM's capability to capture redundant patterns in the test-time scaling paradigm remains underexploited in hybrid speculative decoding, highlighting the need for more refined and dynamic hybrid strategies to fully unlock its potential.
\end{takeaway}

\noindent\textbf{$\bullet$ Time consumption proportions across modules in speculative decoding.} \Figref{ssy0801:time_modules} illustrates the proportion of time consumed by each stage within PIA, EAGLE-3, Recycling, and SAM. During draft generation, the model produces a sequence of tokens based on the current state. The decoding stage processes this draft sequence to compute representations and logits for each position. In the verification phase, the model determines which tokens in the draft are correct by comparing predicted probabilities given by the target model. Finally, the updating stage accepts the verified correct tokens and updates the model state accordingly, preparing for the next iteration. As shown in the \Figref{ssy0801:time_modules}, PIA achieves the lowest proportion of draft generation time, which benefits from its pruning strategy for the n-gram draft tree. In contrast, EAGLE-3 exhibits the highest draft generation time proportion, reaching up to 35.7\%, as it requires invoking an additional draft model that is more time-consuming than n-gram methods. Recycling demonstrates the highest update time proportion among all methods, attributed to its need to dynamically maintain a top-k next token list throughout the generation process. For SAM, the majority of time consumption is concentrated in the target model's decoding phase, which benefits from its advanced suffix automaton data structure. This approach finds the longest suffix that matches the sequence in a suffix automaton at each step of the generation with an average $\mathcal{O}$(1) time complexity, while the linear drafting strategy further reduces computational overhead, significantly minimizing the time spent in other components. 
\begin{takeaway}
    N-gram-based methods, such as SAM, incur lower draft generation time overhead, enabling a greater share of computational resources to be allocated to the decoding phase.
\end{takeaway}

\noindent\textbf{$\bullet$ Analysis of speculative decoding on models with different scales.} In addition to evaluating Qwen3-8B models, we further investigate the effectiveness of speculative decoding methods across different model scales (as shown in Table \ref{ssy0801:different_size}), including the smaller Qwen3-4B and larger models such as Qwen3-14B. Among them, SAM${\text{[EAGLE-3]}}$ achieves the best performance across all model sizes, demonstrating the potential of hybrid speculative decoding. We also observe that EAGLE shows slight performance degradation when accelerating 4B/14B models compared to the 8B scenario. This stems from the training process, as the EAGLE draft models for 4B/14B come from a different team AngelSlim\footnote{\url{https://huggingface.co/AngelSlim}}, highlighting how training-based speculative decoding methods are highly dependent on the draft model's training quality. Benefiting from the increased model size, we find that the SpS for the 14B model shows significant improvement compared to the 8B model. This is primarily because the 14B target model is proportionally much larger relative to the 0.6B draft model than the 8B model is, enabling more effective speculation.

\section{Conclusion}
In this paper, we introduce the first benchmark for speculative decoding in accelerating LLM test-time scaling. Specifically, we evaluate a diverse range of methods, including model-based, training-based, and n-gram-based approaches, under prominent frameworks such as Best-of-N sampling and multi-round thinking. Through extensive experiments, we find that simple N-gram-based methods show strong potential in accelerating test-time scaling by effectively capturing repetitive reasoning patterns. This highlights the value of integrating N-gram-based methods with other categories of approaches, thereby balancing acceleration for both repetitive and diverse reasoning in test-time scaling. We hope this benchmark serves as a foundation for advancing speculative decoding techniques tailored to test-time scaling, fostering more efficient and practical LLM reasoning.

\bibliography{iclr2025_conference}
\bibliographystyle{iclr2025_conference}

\appendix

\end{document}